\documentclass{article}
\usepackage{ijcai24}

\usepackage{times}
\usepackage{soul}
\usepackage{url}
\usepackage[hidelinks]{hyperref}
\usepackage[utf8]{inputenc}
\usepackage[small]{caption}
\usepackage{graphicx}
\usepackage{amsmath}
\usepackage{amsthm}
\usepackage{booktabs}
\usepackage{algorithm}
\usepackage{algorithmic}
\usepackage{xcolor}
\usepackage{colortbl}
\usepackage[switch]{lineno}
\usepackage{subcaption}
\usepackage{amsfonts} 

\definecolor{myred}{RGB}{244,181,180}
\definecolor{myorange}{RGB}{249,218,183}
\definecolor{myyellow}{RGB}{255,255,187}
\newcommand{\topone}{\colorbox{myred}}
\newcommand{\toptwo}{\colorbox{myorange}}
\newcommand{\topthree}{\colorbox{myyellow}}

\title{ClothPPO: A Proximal Policy Optimization Enhancing Framework for Robotic Cloth Manipulation with Observation-Aligned Action Spaces}

\author{
Libing Yang$^1$
\and
Yang Li$^1$\footnote{corresponding auther}\And
Long Chen$^{2}$
\affiliations
$^1$East China Normal University\\
$^2$The Hong Kong University of Science and Technology
\emails
51255901139@stu.ecnu.edu.cn,
yli@cs.ecnu.edu.cn,
longchen@ust.hk
}

\begin{document}

\maketitle
\begin{abstract}
% my version 
Vision-based robotic cloth unfolding has made great progress recently. However, prior works predominantly rely on value learning and have not fully explored policy-based techniques.
Recently, the success of reinforcement learning on the large language model has shown that the policy gradient algorithm can enhance policy with huge action space.
In this paper, we introduce ClothPPO, a framework that employs a policy gradient algorithm based on actor-critic architecture to enhance a pre-trained model with huge $10^6$ action spaces aligned with observation in the task of unfolding clothes. 
To this end, we redefine the cloth manipulation problem as a partially observable Markov decision process.
% Inspired by advances in reinforcement learning on the large language model,
A supervised pre-training stage is employed to train a baseline model of our policy.
In the second stage, the Proximal Policy Optimization (PPO) is utilized to guide the supervised model within the observation-aligned action space.
By optimizing and updating the strategy, our proposed method increases the garment's surface area for cloth unfolding under the soft-body manipulation task. 
Experimental results show that our proposed framework can further improve the unfolding performance of other state-of-the-art methods.
% The experiment shows that our proposed framework can successfully improve the unfolding performance of other methods.
% Robotic cloth manipulation can relieve humans from mundane household chores. 
% Yet, prior works predominantly rely on value learning and have not fully explored policy-based techniques. 
% the disparity between high-dimensional visual observations and low-dimensional action commands. 
% Although policy gradient methods are successful in certain applications, they struggle with the complexity of accurate cloth manipulation due to the disparity between low-dimensional action commands and high-dimensional visual observations.
% % teacher version
% Although robotic cloth unfolding has made great progress, learning with a large action space in manipulation is still a challenge and difficult problem. Recently, the success of reinforcement learning on the large language model has shown that the policy gradient algorithm can successfully solve the problem of huge action space in the supervised fine-tuning model. In this paper, we propose a proximal policy optimization-based approach to enhance the manipulator in cloth unfolding. To this end, an observation-aligned action space is defined to alleviate the difficulties in sampling action. By optimizing and updating the strategy, our proposed method increases the garment's surface area for clothing unfolding under the soft-body manipulation task. Experimental results show that our proposed framework can further improve the unfolding performance of other state-of-the-art methods.

\end{abstract}

\section{Introduction}
With the continuous technological advancements in the field of embodied intelligence~\cite{gupta2021embodied}, emerging application scenarios for robots are gradually coming to the fore, such as service robots.  
The manipulation of flexible objects is a fundamental and common skill in human domestic labor, including activities such as unfolding and tidying. A robot capable of manipulating clothes could alleviate the burden of household chores and significantly enhance the practicality of robots in our lives.

Compared to rigid body manipulation~\cite{zeng2020transporter}, predicting morphological changes in fabrics through visual inputs is very difficult due to their high dimensionality and complex dynamics~\cite{seita2021learning,lin2022learning} in cloth manipulation.
The reliance on supervised learning for such tasks~\cite{xiong2023robotube} demands an extensive collection of labeled training data, which can be a time-intensive and arduous process.~\cite{matas2018sim} proposes to use self-supervised learning to deal with the manual labeling problem. 
However, supervised learning models often exhibit limited generalization capabilities~\cite{nair2017combining}. They tend to excel in conditions similar to their training environments but may falter in new or unforeseen scenarios.
This characteristic becomes particularly problematic in robotic manipulation. 
The emergence of Deep Reinforcement Learning (DRL)~\cite{franccois2018introduction} offers a new paradigm to achieve more generality and adaptability. 
Previous methods~\cite{ha2021flingbot,canberk2022clothfunnels,avigal2022speedfolding} employ a value function to predict the unfolded coverage in the task of folding clothes and achieve very impressive results. 
By employing supervised learning on the coverage area values directly in the observation~\cite{wu2020spatial}, these methods successfully build the robot's ability to increase the performance of unfolding garments. 
However, these methods inherit self-supervision from the DRL framework but do not use multi-step reward, still highly rely on the accuracy of the value function. This makes these methods inherit the problems of supervised learning.

On the other hand, policy-based reinforcement methods~\cite{chen2023daxbench} have shown mediocre performance in practice. 
Traditional action commands in policy-based DRL are typically represented by low-dimensional discrete or continuous vectors. 
For instance, robot actions are often controlled by specifying directions (forward, backward, left, right) and movement distances. 
This results in an inherently discrete low-dimensional action space, compared with the high-dimensional nature of visual input space. 
~\cite{wu2020learning} employ CNN-based networks and MLPs to map the visual inputs to this low-dimensional action space.
As the essential complexity of the garment's dynamics, this mapping potentially leads to significant information loss, which hinders the learning of policy. However, the dual-pose action directly within the pixel space meets another problem an exponential escalation in the number of feasible actions. The pixels in the image as a potential point of action.
Considering an image dimension of 256, the resulting potential action choices would exponentially rise to the magnitude of \( 10^{10} \), which is impossible to solve.

In this paper, we propose ClothPPO to fully take advantage of reinforcement learning for cloth unfolding tasks. 
To this end, an observation alignment policy is introduced to map the image directly to a high-dimension action distribution similar to the spatial action maps~\cite{wu2020spatial}.
To address the information loss from images to low-dimensional actions and directly connect image pixels with robot actions, we first to use spatial action maps in policy-based RL to model policy. 
This makes our proposed policy have enough capability to capture complex dynamics of fabric changes during manipulation.
In addition, inspired by the success of reinforcement learning from human feedback~\cite{ouyang2022training} on large language model, a two-stage learning framework with PPO~\cite{schulman2017proximal} is proposed to solve the large action space learning problem~\cite{Ramamurthy2022IsRL} in policy-based RL.
At first, we offline pre-train a UNet-based policy in a self-supervised learning fashion.
Then an actor-critic architecture~\cite{schulman2017proximal} is employed with PPO to further online fine-tune the proposed policy.
% To the best of our knowledge, our proposed approach is the first offline pre-training to online fine-tuning for cloth manipulation tasks.
% For PPO training, we implement Cloth Action Gym with our huge observation alignment action spaces based on the Gym interface~\cite{brockman2016openai} on an existing Pyflex environment~\cite{canberk2022clothfunnels}. 
% It contributes to the community by promoting further exploration and research in robotic manipulation and RL.
% Experimental results show that our proposed method enhances the pre-train model in cloth unfolding tasks with pick-place primitive and can improve other pick-and-place primitives methods~\cite{canberk2022clothfunnels} from 75\% to 82\% in unfolding coverage.

We have the following three contributions. 
1)Our method generates actions by sampling from a distribution-modeled policy, encouraging thorough exploration of the action space without relying on accurate estimations. Our work is the first to use a policy-based RL approach to directly select robot actions in the huge pixel space aligned with observation.
2) The vast action space makes the exploration process of the RL inefficient and difficult to learn. Inspired by the training process of ChatGPT, we bridge value-based and policy-based approaches, using the value-based method for initialization and employing PPO to enhance the pre-trained model's effectiveness in addressing the issue of large action spaces in RL. 
3) We apply this method to the complex task of robotic cloth unfolding which achieves SOTA results on this difficult practical application problem, and contribute an environment based on our method to the robotics community.

\section{Related Work}
% \subsection{Cloth manipulation}
Research in the domain of flexible cloth manipulation has extensively focused on vision-based methodologies to orchestrate manipulative actions for an array of tasks, encompassing folding and unfolding. In the nascent stages of robotic garment folding, prevalent practices entailed initially unfurling the cloth into predefined, antecedently known states, subsequently engaging in the automated folding via procedural or rudimentary rule-based heuristic techniques~\cite{triantafyllou2016geometric}.
% add extra
The perception module mainly uses CNN~\cite{avigal2022speedfolding,hu2023disassembling}, Transformer~\cite{mo2022foldsformer,chen2024life} and other network structures.

In robotics cloth manipulation, the primitives used for manipulation are categorized into two types: quasi-static and dynamic. Previous work has actively explored the determination of parameters for predefined operational primitives ~\cite{mulero2023qdp}. Dynamic methods calculate high-speed dynamic throwing actions directly from overhead images, which can smooth a garment to 80\% coverage within an average of three actions~\cite{ha2021flingbot}. 

Current mainstream approaches are divided into model-based learning, which models the environment's dynamics while neglecting policy learning, and model-free learning, which derives cloth manipulation strategies from images through supervised or reinforcement learning. 
The latter faces challenges in achieving high robustness due to ambiguities in 2D observational data, insufficiently flexible kinematic structures, and complex environmental interactions.
Some papers address this issue by learning from demonstrations~\cite{xiong2023robotube}. Demonstrations can be obtained from scripted action sequences based on heuristic experts~\cite{doumanoglou2016folding} or cloth descriptors~\cite{ganapathi2021learning}. ~\cite{avigal2022speedfolding} engage in self-supervised learning augmented by minimal expert demonstrations to execute cloth smoothing and folding with an assortment of manipulation primitives, achieving a folding success rate as high as 93\% within 120 seconds from a randomized initial configuration. However, this method is trained directly in the real world, has weak generalization, requires costly manual demonstration annotations, and has high training costs. 
Another learning approach is model-free reinforcement learning,~\cite{canberk2022clothfunnels} introduces a self-supervised reward-based approach for learning cloth normalization and alignment, diminishing the complexity of specific task-oriented downstream policies. However, due to the cloth's potential for various wrinkle structures, it is challenging to generalize to different target situations. 
There is also work primarily based on models, suggesting methods for modeling the dynamics of flexible objects for manipulation.~\cite{lin2022learning} employing a particle-based GNN dynamic model informed by point cloud observations. Another study~\cite{huang2023self} proposes estimating the real-world cloth 3D mesh under action conditions. Yet, learning accurate 3D representations of real flexible objects within an end-to-end framework remains difficult, the effectiveness of such methods is heavily reliant on the performance of 3D perceptual dynamics modeling.

\section{Problem Formulation}
\begin{figure*}[htbp]
    \centering
    \includegraphics[clip, trim=0cm 0cm 0cm 0cm, width=0.8\textwidth]{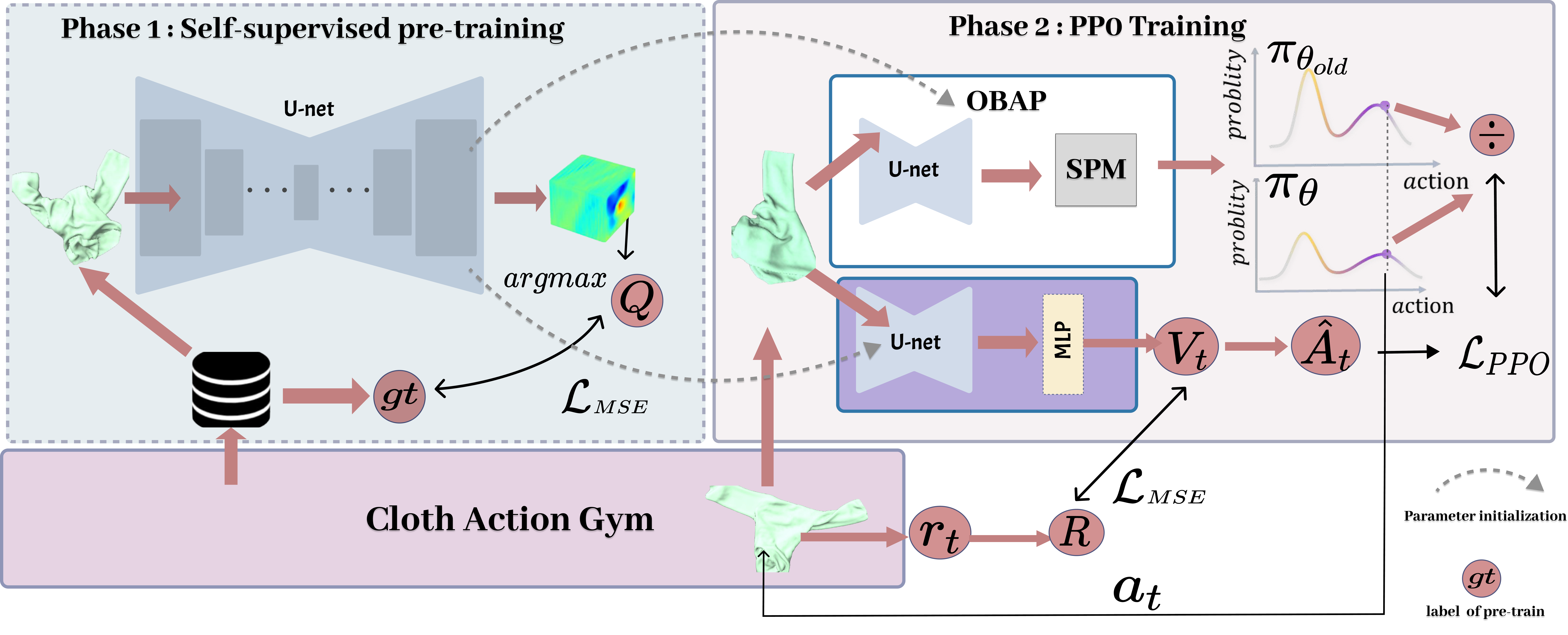}
    \vspace{-0.3cm} % 调整标题和图片之间的间隔
    \caption{\textbf{Overview of ClothPPO.} The first phase  \protect\cite{canberk2022clothfunnels} involves self-supervised pre-training, which uses data from repeated actions in the environment to collect labels. The model estimates canonicalized alignment grasping label and selects the maximum estimated value to output the action. We introduce a long-term reward mechanism to improve the model's performance in goal-oriented tasks in the PPO training phase.}
    \label{fig: ClothPPO}
\end{figure*}
\subsection{Markov Decision Process}
The task of cloth unfolding presents a complex challenge, where robots must interact with fabric over discrete time steps $t$ to maximize coverage area. This necessitates devising an effective manipulation strategy to explore the problem of learning robotic actions \( a_t \) from visual observations \( o_t \). 
A key difficulty lies in the inability to fully observe the cloth's state in real-world scenarios. It relies on partial visual inputs, creating a gap between perception and complete state awareness. Therefore, we model cloth manipulation as a Partially Observable Markov Decision Process (POMDP)~\cite{sondik1971optimal}.
% It can be formally described as $(S, A, O, T, R, \gamma)$, where each element represents a different aspect of the process. 
% \textbf{State Space} \(S = \{s_1, s_2, \dots, s_t\}\): This is the set of all possible states in the environment. In the context of cloth unfolding, each state represents a different configuration of the cloth.
% \textbf{Action Space} \( A = \{a_1, a_2, \dots, a_t\} \): This set includes all the actions the agent can take.
% \textbf{Observation Space} \( O = \{o_1, o_2, \dots, o_t\} \): These are the possible observations the agent can make about the state of the environment. Due to partial observability, these observations don't provide complete information about the state.
% \textbf{Transition Function} \( T(s_t, a_t) = P(s_{t+1}| s_t, a_t) \): This function defines the probability of transitioning from one state to another, given an action.
% \textbf{Reward Function} \( R(s, a) =  \): This function gives the expected reward received after taking an action in a given state.
% \textbf{Discount Factor} \( 0 \leq \gamma \leq 1 \): A value between 0 and 1 that determines the importance of future rewards. A lower value places more emphasis on immediate rewards.
A POMDP is defined by the tuple $(\mathcal{S}, \mathcal{A}, \mathcal{O}, P, O, R, \gamma)$, where
$\mathcal{S}$ denotes the state space of all possible configurations of the cloth.
$\mathcal{A}$ represents the action space, comprising all the manipulative actions a robot can perform on the cloth.
$\mathcal{O}$ is the observation space, reflecting the visual observations $o_t$ available to the robot at time $t$.
$P(s' | s, a)$ is the state transition probability function that provides the probability of moving to state $s'$ from state $s$ after taking action $a_t$.
$O(o | s', a)$ is the observation function that gives the probability of receiving an observation $o$ after taking an action $a_t$ and ending up in state $s'$.
$R(s, a)$ is the reward function that assigns a numeric reward for taking an action $a_t$ in state $s$, guiding the robot to maximize the cloth's coverage area.
$\gamma \in [0, 1]$ is the discount factor that values the importance of immediate versus future rewards. Suppose the strategy parameter is $\theta$ and our optimization objective is to maximize the average return. A reward is $r_t = r(o_t, a_t,o_{t+1}) $. Given the sequence tuples of $(o_t, a_t, r_t)$.  The optimization objective can be expressed as $J(\theta) = E[R]$, where $J(\theta)$ is the objective function of the strategy and $R = \sum_{t=0}^{T} \gamma^{t}r_t$:
\begin{equation}
    \pi_\theta  = \underset{\pi}{\mathrm{argmax}} 
 \mathbb{E}_{\pi}\left[\sum_{t=0}^{T} \gamma^{t} R(o_t, a_t)\right],
 \label{eq: return}
\end{equation}
It is to find a policy $\pi_\theta(a_t|o_t)$ that maximizes the expected sum of discounted future rewards. This involves calculating an expected value that balances immediate and future rewards. To increase the probabilities of taking actions in high-reward trajectories.
% and to decrease the probabilities in low-reward trajectories. 
In this way, during the training process, the policy gradually adjusts its parameters to improve its performance.
\subsection{Definition of Observation and Action}
% We define the action space of the above PDMDP as a multivariate discrete action space as $\mathcal{A}$, where each element represents a different action.
% \begin{equation}
%      \mathcal{A} = x \times y \times r \times l
% \end{equation}
% where each element represents a different dimension of the action space: \(x\) represents all possible actions along the x-axis, \(y\) represents all possible actions along the y-axis,
% \(r\) is a set composed of angles \(\{\theta_1, \theta_2, \dots, \theta_n\}\), all of which are within the range \([-180, 180]\) degrees. This range is equally divided into \(n\) parts, each angular interval is the total range divided by \(n\). That is, the interval between each angle is \(\Delta\theta = \frac{360}{n}\) degrees.
% And \(l\) represents all possible actions along the length axis. If the length axis is within the range \([a, b]\), and we divide it into \(m\) equal parts, each segment length is \(\Delta l = \frac{b-a}{m}\).

In the task of clothing unfolding, the agent is assigned to grasp the points on the cloth and unfold it into a flat state. This task can be accomplished through a series of dual-pose motion primitives, such as pick and place.
Given the original header observation, policy decides the optimal actions $a_t = (a_{p_1}, a_{p_2}) \in \mathcal{A}$. Observation is an RGBD image $o_t \in \mathbb{R}^{W \times H \times C}$, where H and W are the height and width dimensions of the image. 
The function that maps observations to actions is defined as:
\begin{equation}
    \pi_{\theta}(a_{p1}, a_{p2} | o_t) = p(a_{p1} | o_t) \cdot p(a_{p2} | o_t, a_{p1}),
    \label{eq: action 2pose}
  \end{equation}
% where \( \mathcal{A} \) denotes a multi-discrete action space that includes all possible actions. 
where \(p(a_{p_1}|o_t)\) and \(p(a_{p_2}|o_t, a_{p_1})\) are the probabilities of these actions given the observation \(o_t\).
% Subactions \( a_{{p_1}} \) and \( a_{{p_2}} \), which are defined within either the SE(2) or SE(3) space depending on the task requirements and the available degrees of freedom of the end effector. 
Since observations can be projected into a 2D-pixel space, with the third dimension held at a fixed depth, it treats this as a two-dimensional action. This constrains \( a_{{p_1}} \)  and  \( a_{{p_2}} \) to two points in pixel coordinate.

The dual-pose joint action model, which incorporates both direct and visual input modalities, is formalized as a multivariate discrete scalar action space. This space comprises four variables amalgamated within the pixel space. However, the two-step operational action space involved in making decisions directly at the image pixel level is huge.
% By using Spatial Action Maps, a policy network can learn to output actions directly without the need for an intermediate representation. This direct mapping from observation to action can simplify the policy network and reduce the chance of errors in translation. Spatial Action Map aligns input and action spaces, we map these scalars into grasping values with constant scale and rotation, representing a starting point and a line segment associated with scale and rotation. The dual-pose action is also defined by four discrete scalars but using Spatial Action Maps as a representation of dual-pose actions effectively reduces the dimensionality of the action from a larger two-step pixel space action to a one-step pixel action space.

\section{Method}
As shown in Fig.\ref{fig: ClothPPO}, the training process of our proposed method consists of two phases, our contributions focus on the second phase:

\paragraph{Self Supervised Pre-training.}
The model is founded on self-supervised principles and adopts a regression methodology for individual pixel values $Q$ using Mean Squared Error (MSE) loss. However, its inherently step-by-step prediction approach presents a key limitation: a noticeable absence of long-term guidance. 

% During this phase, the model engages in self-supervised learning to process partially observable observations, denoted as \( O \). Given the agent's limited access to environmental information, it cannot fully determine the internal state \( S \). Consequently, the agent must deduce the state \( S \) from the observations. The model cultivates a state representation \( S \) from these observations using self-supervision and regresses the value of individual pixels through an MSE loss.
% \begin{equation}
%     \theta  = \underset{\pi}{\mathrm{argmax}} \mathbb{E}_{\pi}\left[\sum_{t=0}^{T} \gamma^{t} R(s_t, a_t)\right]
% \end{equation}

\paragraph{PPO Training.}
In the second phase, we use the PPO to fine-tune the pre-trained model. In this phase, we introduce a long-term reward mechanism to improve the model's performance in goal-oriented tasks. The PPO iteratively optimizes the model's policy to better adapt to the goal task and guides the model's learning process through reward signals. With these two phases of training, our goal is to enable the model to generate excellent grasping actions in grasping tasks and show higher performance in goal-oriented tasks. This two-stage training strategy improves the generalization ability and task adaptability of the model.

\subsection{Observation Alignment Policy}
% The output is a 3D Affordances Value Spatial Action Map \( C \times H \times W \), where each pixel corresponds to a unique action. These actions are ascertained by the pixel's index within the 3D value maps, with \( H \times W \) determining the point's location and \( C \) representing the choices of segment rotation and scaling. Consequently, a single pixel's value is indicative of \( Q(S_t, A_t) \), and operations are selected through an \( \text{argmax} \) strategy, optimizing action selection.

\paragraph{Redefine Action to Reduce the Action Space.}As formulated Eq.~\ref{eq: action 2pose}, the dual-pose joint action comprises four variables amalgamated within the pixel space. 
We redefine the action space of the above PDMDP as another multivariate discrete action space $\mathcal{A}$,  where each element represents a different action. Just like what common RL algorithms do before, re-describe the joint action using the starting point and the line segment starting from the starting point. Effectively reducing the dimensionality of the action from a larger two-step pixel space action to a one-step pixel \( 10^{6} \) action space, at the same time, the correlation between the two poses is ensured. The action at time \(t\) can be presented as
\begin{equation}
     a_t = (x, y, \phi, d) =  (x, y, z) \in \mathcal{A},
\end{equation}
    where each element represents a different dimension of the action space, \(x\), \(y\) represents the set of action positions in pixel coordinate, \(n\) is the number of a set of angles \(\Phi=\{\phi_i | \phi_i = \frac{360^\circ*i}{n}, i=1,2,...,n\}\). \(m\) represents the number of a set of movement distances \(D=\{d_i | d_i = \frac{b-a}{m}*i+a, i=1,2,...,m\}\), where \(b, a\) represent the max and min movement distance. \(z \in \Phi \times D\) denotes the combination of an angle and a movement distance. 
    
\begin{figure}[t]
    \centering
    \includegraphics[clip, trim=0cm 0cm 0cm 0cm, width=0.4\textwidth]{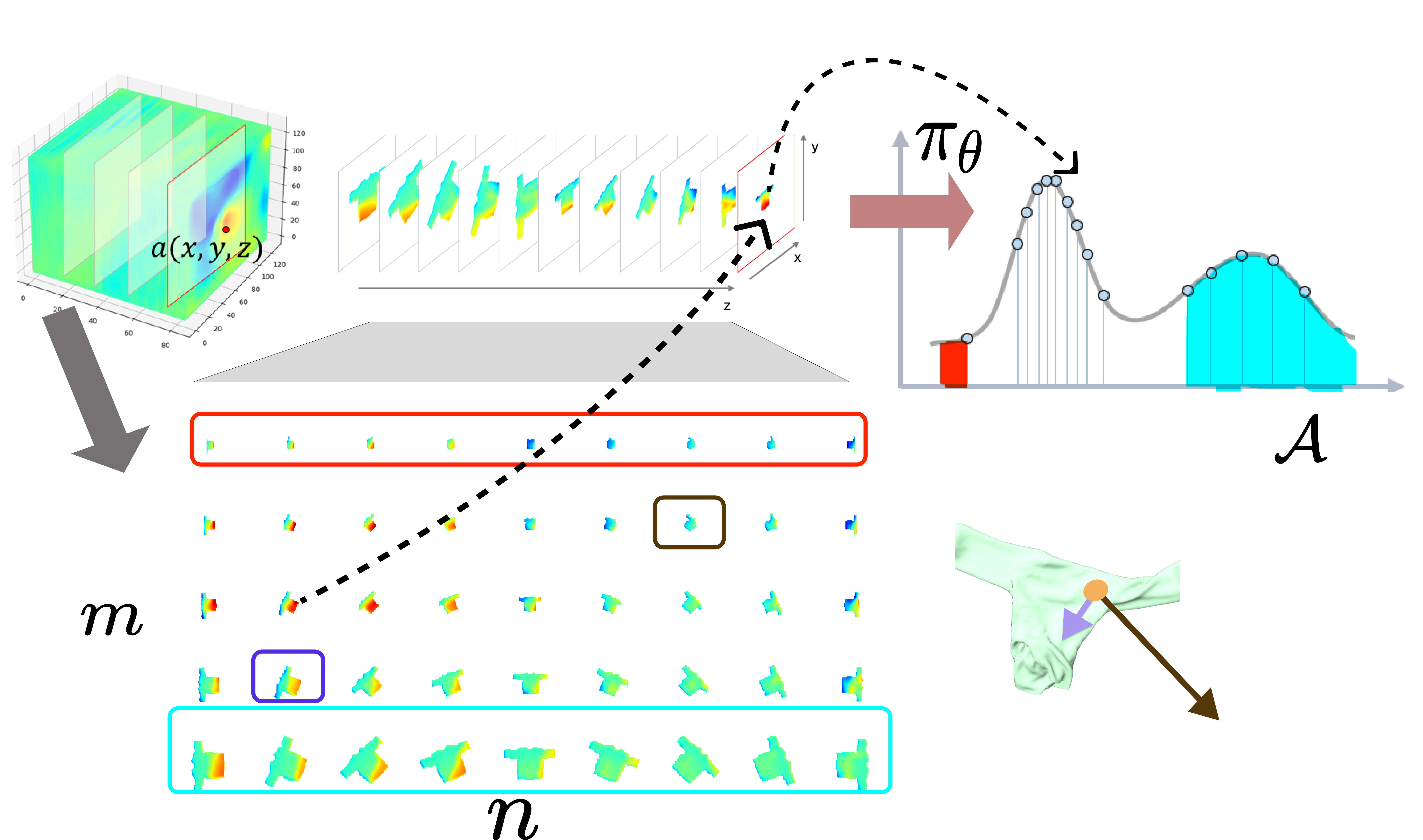}
    \vspace{-0.2cm} % 调整标题和图片之间的间隔
    \caption{SPM: Our action spaces and sampling action using spatial policy maps. The series of smaller maps to the bottom are different slices of the spatial policy maps, each representing a layer at a different scale and rotation. The masks applied to each layer serve to filter out invalid actions those that would result in the robot's end-effector interacting with space or areas beyond the cloth. The variation in the sizes of the masks corresponds to different scales, affecting the size and granularity of the actions that can be sampled.}
    \label{fig: SPM}
\end{figure}

The line segments on the clothes explain this definition in Fig.\ref{fig: SPM}. The black and purple line segments represent two different actions.
\paragraph{Rotate and Scale Observation.}
For intuitive understanding, we refer to the setting of spatial action maps~\cite{wu2020spatial} to rotate and scale the observation to align action space with the visual input, $x, y, z $ is the position of the 3D map. The corresponding joint action is determined by the index of the pixel in the 3D map, so a single pixel can correspond to an action. 
Spatial action maps facilitate the network's ability to learn and predict the $Q$ value for each point, guided by the characteristics of the state representation. It is used to predict the value function $Q(s_t, a_t)$ before. 
We use it as spatial policy maps to model the policy itself, which represents the preference for choosing various possible actions in a given state. Instead of explicitly calculating the value function to optimize the strategy indirectly through the value function.
As far as we know, ours is the first work to use it in policy-based RL.

\subsection{Sampling Action from OBAP}
\paragraph{Flatten 3D Policy Maps to Categorical Distribution.}
Spatial policy maps output from the network indicate the probability of each potential action within the environment. To sample an action $a_t$ according to the policy $\pi_{\theta}(\cdot | o_t)$, the multi-dimensional spatial policy maps must be converted into a one-dimensional logits vector, which represents the log probabilities before normalization. This transformation process is referred to as flattening.

When employing a categorical action spaces~\cite{tutz1991sequential}, the output from the network can be transformed into a discrete probability distribution through a softmax operation on the logits vector, denoted as $\mathbf{z}_t$:
\begin{equation}
p(a_t|o_t) = \frac{e^{\mathbf{z}_{t,a_t}}}{\sum_k e^{\mathbf{z}_{t,k}}},
\end{equation}
where $a_t$ is the action taken at time $t$ and $k$ iterates over all possible actions.

\paragraph{Spatial Policy Mask.}
Sampling actions directly on the existing spatial policy maps will lead to many invalid actions, such as empty actions in parts other than clothes. Due to the different rotation and scaling of each layer, the shape of the cloth mask is also different. We have added a different mask to each layer of the spatial policy maps. We use the masked spatial policy maps to filter out actions that will catch the void, and sample actions from the masked spatial policy maps in the PPO phase. For each state observation $o_t$ at time $t$, the applicable action space is obtained by excluding the invalid actions identified by the mask $M_t$. As indicated by the equation below, we have incorporated a mask into the action space on top of PPO to filter out futile actions:
\begin{equation}
    \pi_{\theta}^{*}(a_t | o_t) = 
    \begin{cases}
    \pi_{\theta}(a_t | o_t) & \text{if } M_t(a_t) = 1, \\
    \text{-inf} & \text{if } M_t(a_t) = 0,
    \end{cases}
    \end{equation}
    where $M_t(a_t)$ is the masking function at time $t$ that returns 1 for valid actions and 0 for invalid actions. 

As shown in the red area in Fig.~\ref{fig: SPM}, if the observation is scaled to smaller, from an action perspective, the line segment will be relatively longer, and the movement will be further. From the perspective of observation, there are fewer pixels left after being masked. Then its proportion in the distribution will become smaller after flattening.
In other words, violent action will be a small probability to sample, and vice versa (blue area).

The benefit of this Spatial Policy Mask(SPM) is the probability of actions of different granularities being sampled can be controlled by the size of the mask corresponding to the scaling, thus enabling a more efficient search through the action space for the optimal set of movements. In this way, we get our fully Observable Alignment Policy(OBAP).

% Introduction to a correction through a mask
To account for constraints in the action space, such as the presence of invalid or impractical actions, a mask vector $\mathbf{m}_t$ is applied element-wise to the logits before the softmax operation. This mask has binary entries where 1 indicates a valid action and 0 is an invalid one. The probability of selecting any specific action $a_t$ from the masked action space is given by $\tilde{p}(a_t|o_t)$, which respects the feasibility of actions as determined by the constraints encoded in the mask $\mathbf{m}_t$.
\begin{equation}
\tilde{p}(a_t|o_t) = \frac{e^{\mathbf{z}_{t,a_t} \cdot \mathbf{m}_{t, a_t}}}{\sum_k e^{\mathbf{z}_{t,k} \cdot \mathbf{m}_{t, k}}},
\end{equation}
where the exponentiation and subsequent normalization ensure that only valid actions have non-zero probabilities, effectively sampling actions from a masked categorical distribution.

% We use the SPOM output by the network as the affordance map to guide double-pose actions.
% Sampling actions directly on the existing SPOM will lead to many invalid actions, such as empty actions in parts other than clothes. Due to the different rotation and scaling of each layer, the shape of the cloth mask is also different. We have added a different mask to each layer of the SPOM. We use the masked SPOM to filter out actions that will catch the void, and sample actions from the masked SPOM in the PPO phase. For each state observation $o_t$ at time $t$, the applicable action space is obtained by excluding the invalid actions identified by the SPOM mask $M_t$. As indicated by the equation below, we have incorporated a mask into the action space on top of PPO to filter out futile actions:

% Due to the huge action space caused by pixel space, it is difficult to make decisions directly in pixel space. Using Spatial Action Maps as a representation of dual-pose actions effectively reduces the dimensionality of the action from a larger two-step pixel space action to a one-step pixel action space.
    
% where $M_t(a)$ is the masking function at time $t$ that returns 1 for valid actions and 0 for invalid actions (those leading to NaN in the SPOM). During policy updates with PPO, only actions with non-zero probabilities are considered, effectively incorporating the mask into the action space.

% \subsubsection{Sampling Actions from a Masked Categorical Distribution}
\subsection{Proximal Policy Optimization Finetuning}
% Policy gradient reinforcement learning algorithms have been widely applied across various domains. However, prior research indicates that the direct application of Proximal Policy Optimization in cloth unfolding tasks yields suboptimal performance, likely due to the complex dynamics of deformable objects. Inspired by the recent successes achieved with large-scale models in RLHF, we have amalgamated policy gradient techniques with supervised learning approaches. 
By employing the PPO~\cite{schulman2017proximal}, we have augmented the efficacy of the supervised policy in cloth manipulation tasks. 

Throughout the training, an action from the action space is sampled at each timestep. Specifically, we sample an action $a_t$ according to the policy $\pi_{\theta}(a_t | o_t)$ derived from the current model parameters $\theta$.
% Throughout the training, we sample an action from pixel positions outside the mask. In testing, we select actions using the argmax strategy. Given a $\theta$ parameterized policy $\pi_{\theta}$ and a set of rollouts, PPO updates the policy as follows:
The core of the PPO update is the clipped surrogate objective, which is used to modify the parameters $\theta$ of the policy network to maximize the expected return. Given a set of trajectories (rollouts), collected under the current policy $\pi_{\theta_{old}}$, PPO updates the policy parameters by optimizing the following loss function:
% PPO clipped objective function
\begin{equation}
    L^{PPO}(\theta) = \mathbb{E}\left[
    \min\left(
    p_t(\theta) \hat{A}_t, \text{clip}\left(p_t(\theta), 1 - \epsilon, 1 + \epsilon\right) \hat{A}_t
    \right)
    \right],
    \label{eq: ppo loss}
    \end{equation}
    where $\hat{A}_t = r_t + \gamma V_t(o_{t+1}) - V_t(o_t)$, is an estimator of the advantage function at timestep $t$, $p_t(\theta)$ denotes the probability ratio $p_t(\theta) = \frac{\pi_{\theta}(a_t | o_t)}{\pi_{\theta_{old}}(a_t | o_t)}$, and $\epsilon$ is a hyperparameter, typically a small constant like 0.1 or 0.2, that defines the clipping range to avoid overly large policy updates. The advantage function estimator $\hat{A}_t$ quantifies the benefit of taking action $a_t$ in $o_t$ over the expected value of the policy at that time. It is calculated using a value function $V(o_t)$. $\mathbb{E}[\cdot]$ is taken over the distribution of actions and states encountered under the current policy $\pi_{\theta_{old}}$. 

% This objective seeks to improve the policy by increasing the probability of actions that lead to higher than expected returns, while penalizing the probability increase for actions that result in lower than expected returns. The clipping function serves as a guardrail, ensuring that the updates are kept within reasonable bounds, thus promoting gradual policy improvement and maintaining training stability.

% Generalized Advantage Estimation return

% $TD(\lambda)$ return estimation: PPO implements the return target as returns = advantages + values.

\paragraph{Model Architecture.}
As shown in Fig.~\ref{fig: ClothPPO}, the PPO phase fine-tuned the OBAP in simulation via an actor-critic architecture~\cite{schulman2017proximal}.

\textbf{Actor.} OBAP that selects actions based on the current observation. The goal of the actor is to maximize the cumulative reward. OBAP accepts RGBD observations \(o_t\) as input and outputs per-step action \(a_t\). OBAP gets the next observation \(o_t\) by interacting with the environment with \(a_t\) and optimizes its parameters based on the multi-step reward to enhance the policy's performance. 
In the initial phase of PPO training, OBAP is a copy of the pre-train model. As the PPO finetunes proceed, the parameters of OBAP are iteratively updated through the policy gradient (see Eq.~\ref{eq: ppo loss}).

\textbf{Critic.} The critic is the value function $V(o_t)$ (see Eq.~\ref{eq: ppo loss}) estimator that evaluates the performance of OBAP. The label of the critic is calculated by muti-step rewards $R(o_t, a_t)$ (see Eq.\ref{eq: return}). The critic accepts \(o_t\) as input and outputs an estimate of the corresponding $R(o_t, a_t)$. Critic obtains reward signals by interacting with the environment and uses these signals to update the parameters of the value function. The network consists of a pre-trained U-Net combined one-layer MLP, that is updated using MSE loss for optimization.

\subsection{Training Details}

\paragraph{Reward Design.} In addressing the cloth unfolding task, our objective is to maximize the cloth's coverage over the workspace, referred to as Coverage. To effectively guide the agent toward achieving higher coverage, we formulate a reward function that integrates both sparse and dense rewards. The reward at each time step $r_t$, is determined by assessing changes in coverage. This change is calculated as the ratio of the difference in current and previous coverage to the maximum possible coverage. The agent will receive a sparse reward of $r_t = -1$ if the coverage decreases, and a dense reward of $r_t = \Delta_{Coverage} \times 20$ if the coverage increases. Additionally, achieving a Coverage greater than 0.9 triggers a sparse reward of 5, further motivating the agent towards optimal cloth manipulation.
\paragraph{Reward Scaling.}
Given the presence of abrupt, sparse rewards in the reward distribution, the network training can experience instability due to excessively large reward magnitudes. To mitigate such instability, it is essential to scale the rewards to act as normalization, thereby stabilizing the training process.
Given the presence of abrupt, sparse rewards in the reward distribution, overly large numbers can cause instability during network training. Therefore, we have scaled the rewards to act as normalization, stabilizing the training. The scale is realized by dividing the reward $r_t$ by the standard deviation of the reward sequence $RS$. This is mathematically reflected in the following equation:
\begin{equation}
    \tilde{r}_t = \frac{r_t}{\sigma(RS)},
    \end{equation}
    where $\tilde{r}_t$ represents the normalized reward at time $t$, and $\sigma(RS)$ denotes the standard deviation of the reward sequence $RS$. By applying this operation, the rewards retain their relative differences in value but are expressed in measures of standard deviations from the mean of the reward distribution.

\section{Experiments Settings}
\paragraph{Environment and Tasks.} Referring to the setting of~\cite{canberk2022clothfunnels}, the simulation environment is built on top of PyFleX~\cite{li2018learning} bindings with Nvidia FleX bindings provided by SoftGym~\cite{corl2020softgym}. We randomly sample the initial state of the cloth from a subset of Cloth3D~\cite{bertiche2020cloth3d}, which contains filtered meshes of various garments. 
We only trained on the T-shirt task and tested on five different tasks of 200 wrinkled clothing with low coverage. To evaluate our policy, we load tasks from the testing task datasets and then run the policy for 10 steps or until the cloth is out of observation or 10 steps end. 
\paragraph{Metrics.} In the simulation, the workspace is discretized into a grid. Each cell on the grid is either covered by cloth or not. We define a coverage metric $C_{sim}$, which is the sum of the areas of all grid cells covered by the cloth: $C_{sim} = \sum_{i=1}^{N} A_i$, where $N$ is the number of cells covered by the cloth and $A_i$ is the area of the $i$-th cell. To evaluate the cloth manipulation performance, we compute the coverage percentage $C_{pct}$, which is the ratio of $C_{sim}$ to the total area $A_{flat}$ designated for the task, multiplied by 100:
$ C_{pct} = \frac{C_{sim}}{A_{flat}} \times 100\%.$
Final coverage measures the coverage at the end of an episode and delta coverage is final coverage minus initial coverage. Percent positive is the proportion of actions that increased coverage.
% In the simulation, the cloth is modeled as a grid of particles, and the coverage metric quantifies the area effectively covered by the cloth within the workspace. Let \( \textbf{pos} \) denote the array of particle positions within the environment. The array \( \textbf{pos} \) is structured such that each particle's position is given by its coordinates in the simulation space.

% The coverage at any time is calculated by summing up the areas of all grid cells that are covered by the cloth. 
% \begin{equation}
%     Coverage = \sum_{i=1}^{N} a_i,
% \end{equation}
% where \( N \) is the number of grid cells covered by the particles, and \( a_i \) represents the area of the \( i \)-th cell. The areas influenced by the cloth's particles are marked within the discretized grid, and the total coverage is the sum of these marked areas.

% To measure the performance of the cloth manipulation task, we calculate the coverage percentage, which is the ratio of the cloth-covered area to the total area designated for the task. The coverage percentage is defined as:

% \begin{equation}
%     Coverage = \frac{Coverage_{current}}{Coverage_{flatten}},
% \end{equation}

% where $A_{flatten}$ is the total area that the cloth should cover when fully spread out, as defined by the current task configuration. This provides the coverage, indicating the proportion of the intended area that is actually covered by the cloth in its current state.
\paragraph{Cloth Action Gym.} We build a reinforcement learning environment Cloth Action Gym that is easy to use and extend by wrapping an existing simulation with the OpenAI Gym~\cite{brockman2016openai} API. We also apply Tianshou~\cite{tianshou}'s PPO algorithm to implement the above method. In this way, others can quickly modify the existing task environments and high-level action primitives to suit their research goals, and use any reinforcement learning frameworks that are compatible with our environment.

\paragraph{Baselines.} We compare our method with the following baselines on various categories of clothing:
Flingbot~\cite{ha2021flingbot}, which utilizes the dynamic dual-arm Fling action primitive, leveraging coverage as a label in its value map. We employ the original model and action primitive configurations provided by Flingbot, which is trained on rectangular fabric of varying sizes. Experiments are conducted on multi-category clothing tasks.
PPO means the PPO algorithm without pre-training model.
Cloth Funnels~\cite{canberk2022clothfunnels}, a multi-action primitive method for dual-arm robots using canonicalized-alignment labels that requires shape priors.
Cloth Funnels-P, Cloth Funnels, which solely employs the Pick-Place action in line with our action, is also the pre-trained model of ours ClothPPO.
% In our study, we benchmark our methodology against ClothFunnels~\cite{canberk2022clothfunnels}, a leading-edge technique in cloth manipulation. ClothFunnels implements a spatial action map, allocating actions to pixels based on their maximal Q values. This system ingests a frontal RGB image of the cloth, dimensioned at $H \times W \times 3$, and employs a composite transformation encompassing five scales and 16 rotations. This process yields a diverse set of observations, varying in rotation and scale, from which an output image is derived. This image, consistent in scale and rotation, reflects Q values across the pixel space. The optimal action is identified by pinpointing the pixel with the supreme Q value.

\section{Results}
% \begin{table*}[h!]
%     \centering
%         \caption{Comparison with baseline methods across various clothing categories Final/Best Coverage.}
%     \begin{tabular}{lrrrrr}
%         \toprule
%         Method  & Shirt & Pants & Jumpsuit  & Skirt & Dress \\
%         \midrule
%         Flingbot     & 0.5364/0.7139 & 0.6566/0.7803 & 0.5315/0.7109 & 0.5626/0.6929 & 0.6638/0.7865 \\
%         Cloth Funnels & 0.8516/0.9011 & 0.7081/0.8232 & 0.7313/0.8037 & 0.7799/0.8995 & 0.8242/0.9186 \\
%         Cloth Funnels-P & 0.7308/0.7530 &  0.7177/0.8194 & 0.7376/0.7607 & 0.6970/0.7488  & 0.7634/0.8075 \\ 
%         PPO & - & 0.5228/0.6055 & 0.5024/0.5811 & 0.4907/0.5796 & 0.5254/0.6238 \\
%         ClothPPO     & 0.8024/0.8249 & 0.8058/0.8254 & 0.8304/0.8439 & 0.8574/0.8891 & 0.8735/0.0.9004\\
%         \bottomrule
%     \end{tabular}
%     \label{tab: comparison_with_baselines}
% \end{table*}

\begin{table*}[h!]
    \centering
        \caption{Comparison with baseline methods across various clothing categories Final Coverage$\uparrow$/Delta Coverage$\uparrow$/Percent Positive$\uparrow$.}
        \vspace{-0.35cm}
    \begin{tabular}{l|ccccc}
        \toprule
        \scriptsize{} & \scriptsize{Shirt} & \scriptsize{Pants} & \scriptsize{Jumpsuit} & \scriptsize{Skirt} & \scriptsize{Dress} \\
        \midrule
        \scriptsize{Flingbot} & 
        \scriptsize{53.64 / 00.15 / 51.81} & \scriptsize{65.66 / 01.24 / 51.65} & \scriptsize{53.15 / 00.07 / 49.62} & \scriptsize{56.26 / 00.78 / 52.32} & \scriptsize{66.38 / 01.07 / 51.27} 
        \\
        \scriptsize{PPO} & 
        \scriptsize{50.48 / 07.34 / 52.96} & \scriptsize{52.28 / 01.97 / 52.14} & \scriptsize{50.24 / -0.66 / 51.61} & \scriptsize{49.07 / 14.45 / 53.00} & \scriptsize{52.54 / -18.91 / 53.78} 
        \\
        \scriptsize{Cloth Funnels} & 
        \scriptsize{\topone{86.91}/\topone{40.98}/\topthree{71.10}} & \scriptsize{\topthree{70.81}/\topthree{20.60}/\topthree{54.22}} & \scriptsize{\topthree{73.13}/\topthree{21.45}/\topthree{62.89}} & \scriptsize{\toptwo{77.99}/\toptwo{40.82}/\topthree{56.09}} & \scriptsize{\toptwo{79.15}/\toptwo{33.23}/\topthree{55.40}} 
        \\
        \scriptsize{Cloth Funnels-P} & 
        \scriptsize{\topthree{73.08}/\topthree{29.57}/\toptwo{72.14}} & \scriptsize{\toptwo{71.77}/\toptwo{22.77}/\toptwo{55.62}} & \scriptsize{\toptwo{73.76}/\toptwo{23.03}/\toptwo{72.68}} & \scriptsize{\topthree{69.70}/\topthree{35.44}/\toptwo{66.14}} & \scriptsize{\topthree{76.44}/\topthree{30.43}/\toptwo{68.52}} 
        \\ 
        \scriptsize{ClothPPO} & 
        \scriptsize{\toptwo{80.24}/\toptwo{34.34}/\topone{73.84}} & \scriptsize{\topone{80.58}/\topone{30.38}/\topone{72.14}} & \scriptsize{\topone{83.04}/\topone{28.10}/\topone{77.17}} & \scriptsize{\topone{85.74}/\topone{45.17}/\topone{74.22}} & \scriptsize{\topone{86.56}/\topone{40.39}/\topone{73.36}} 
        \\
        \bottomrule
    \end{tabular}
    \label{tab: comparison_with_baselines}
\end{table*}

% \begin{table}[h!]
%     \centering
%     \begin{tabular}{lcccc}
%         \toprule
%         Method  & Shirt & Pants & Jumpsuit & Average \\
%         \hline
%         Flingbot     & 0.5364/0.7139 & 0.7139 & 0.6210 & 0.6238 \\
%         Cloth Funnels &  & 0.7530 & 0.6789 & 0.7209 \\
%         Cloth Funnels-P & 0.7308/0.7530 & - & - & - \\ % Add specific values if available
%         ClothPPO     & 0.8024/0.8249 & 0.8058/0.8254 & 0.8304/0.8439 & 0.7910 \\
%         \hline
%     \end{tabular}
%     \caption{Comparison with baseline methods across various clothing categories.}
%     \label{tab: comparison_with_baselines}
% \end{table}
% \begin{table}[h!]
%     \centering
%     \begin{tabular}{lrrrrrr}
%         \toprule
%         Method  & Shirt & Pants & Jumpsuit & Skirt & Dress\\
%         \midrule
%         Flingbot     & 0.536 & 0.657& 0.534 & 0.563 & 0.664 \\
%         % Cloth Funnels & 0.851 & - & - \\
%         Pre-trained & 0.731 & - & - & - & -    \\ % Add specific values if available
%         PPO  & & 0.516 & 0.547  & - & - \\  
%         ClothPPO     & 0.803 & 0.806 & 0.830 & 0.8574 & 0.873\\
%         \bottomrule
%     \end{tabular}
%     \caption{Comparison with baseline methods across various clothing categories Final Coverage.}
%     \label{tab: comparison_with_baselines}
% \end{table}
\paragraph{Comparing with Baselines.}
Table.\ref{tab: comparison_with_baselines} shows the performance comparison of various methods applied to five different types of clothing tasks.
Each entry in the table shows three numbers separated by a slash representing three different performance metrics. Apart from the T-shirt Task, ClothPPO has attained superior results across all other metrics. This is because the shape of prior labels related to clothing categories used by Cloth Funnels is particularly prominent in this Task. But our ClothPPO still performs better on other types of cloth tasks than the method that requires the use of additional robotic arms to operate Fling action primitives. The most compelling evidence of its efficacy is the significant improvement it has shown when compared to the pre-training model. This clearly illustrates the effectiveness of our approach. This makes perfect sense considering our rewards are directly tied to enhance these three metrics.
\paragraph{Generalize to Unseen Cloth Types.}
The reward system in ClothFunnels relies on the alignment of cloth shapes, thereby requiring knowledge of the fabric's geometry beforehand. Consequently, optimal outcomes for different garment shapes necessitate distinct training models. However, in practical terms, this approach is highly inefficient.
Conversely, ClothPPO's requisite reward is based on the coverage of the cloth, which is a universal metric applicable to all clothing categories, thereby obviating the need for training individual models for each distinct class of garments.
Our experiments demonstrate that ClothPPO, even when exclusively trained on the Shirt task, exhibits excellent performance and adaptability across multiple types of clothing. Notably, ClothPPO's ability to retain high-performance metrics across varying garment types indicates a substantial capacity for generalization. This is crucial for practical applications wherein robots may be tasked to handle a wide array of garments.

% \begin{figure}[h]
%     \centering
%     % 第一个子图
%     \begin{subfigure}[b]{0.45\textwidth} 
%         \includegraphics[width=0.45\textwidth]{./figures/Figure_logits_pretrain_place.pdf}
%         % \caption{预训练的逻辑分布} % 如果需要子图标题的话
%         \label{fig: logits_pretrain_place}
%         \includegraphics[width=0.45\textwidth]{./figures/Figure_logits_aftertrain_place.pdf}
%         \caption{\( logits \) distribution before and after PPO finetuning. From the left panel, we can infer that the pre-trained model's \( logits \) are relatively well-constrained, with most values falling near zero. The right panel shows the \( logits \) distribution after fine-tuning the model with PPO.  }
%         \label{fig: logits_aftertrain_place}
%     \end{subfigure}

%     \begin{subfigure}[b]{0.45\textwidth} 
%         \includegraphics[width=\textwidth]{./figures/Figure_pretrain_place.pdf}
%     \end{subfigure}
%     \begin{subfigure}[b]{0.45\textwidth} 
%         \includegraphics[width=\textwidth]{./figures/Figure_aftertrain_place.png}
%     \end{subfigure}
%     \caption{Visualizations of before and after PPO finetuning}
%     \label{fig: visualizations of after ppo train}
% \end{figure}

\paragraph{Comparing Performance Before and After PPO Training.} 
Table.~\ref{tab: vis_of_before_and_after_ppo} demonstrates a comparison of performance on a consistent task set by both the pre-training model and ClothPPO. The presence of robotic grippers highlighted by red lines, suggests an interaction between the robotic manipulator and the cloth. Within the same task, ClothPPO displays a more significant improvement in coverage at every step. In the final stage of the pre-train model, the coverage stagnates due to the action being void. The pre-training model approaches reliance on the $argmax$ action selection strategy, thereby limiting its exploration of other potential actions. In contrast, ClothPPO manifests a broader scope for action exploration.

\begin{table*}[htbp]
    \centering
    \vspace{-0.2cm} % 调整标题和图片之间的间隔
    \begin{tabular}{cl}
        % Row 1
        \rotatebox[origin=c]{90}{\parbox[c]{1cm}{\tiny{ClothPPO}}}
        \hspace{-0.5cm}
        &
        \begin{minipage}{.9\textwidth}
            \centering
            \includegraphics[clip, trim=0cm 31.5cm 0cm 0cm, width=\linewidth]{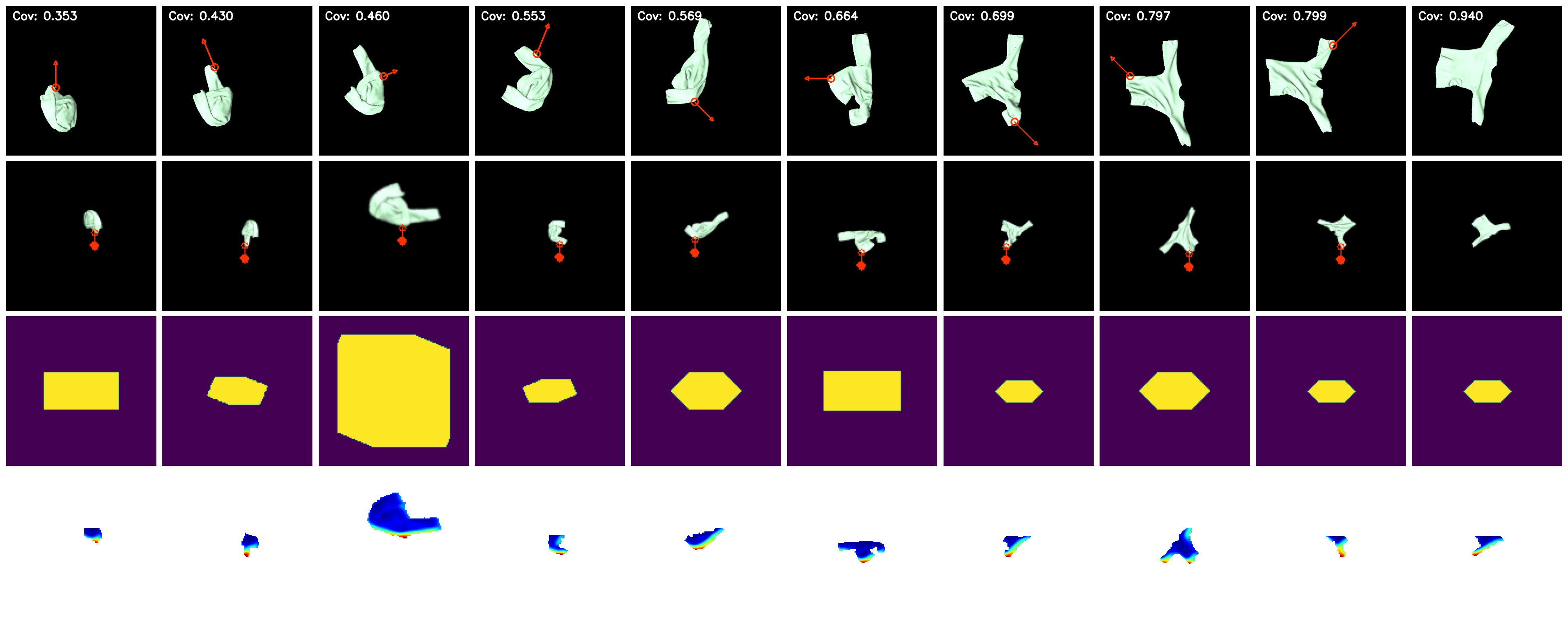}
        \end{minipage} \\
        % Row 2
        \rotatebox[origin=c]{90}{\parbox[c]{1cm}{\tiny{Pretrain}}} 
        \hspace{-0.5cm}
        &
        \begin{minipage}{.9\textwidth} 
            \centering
            \includegraphics[clip, trim=0cm 31.5cm 0cm 0cm, width=\linewidth]{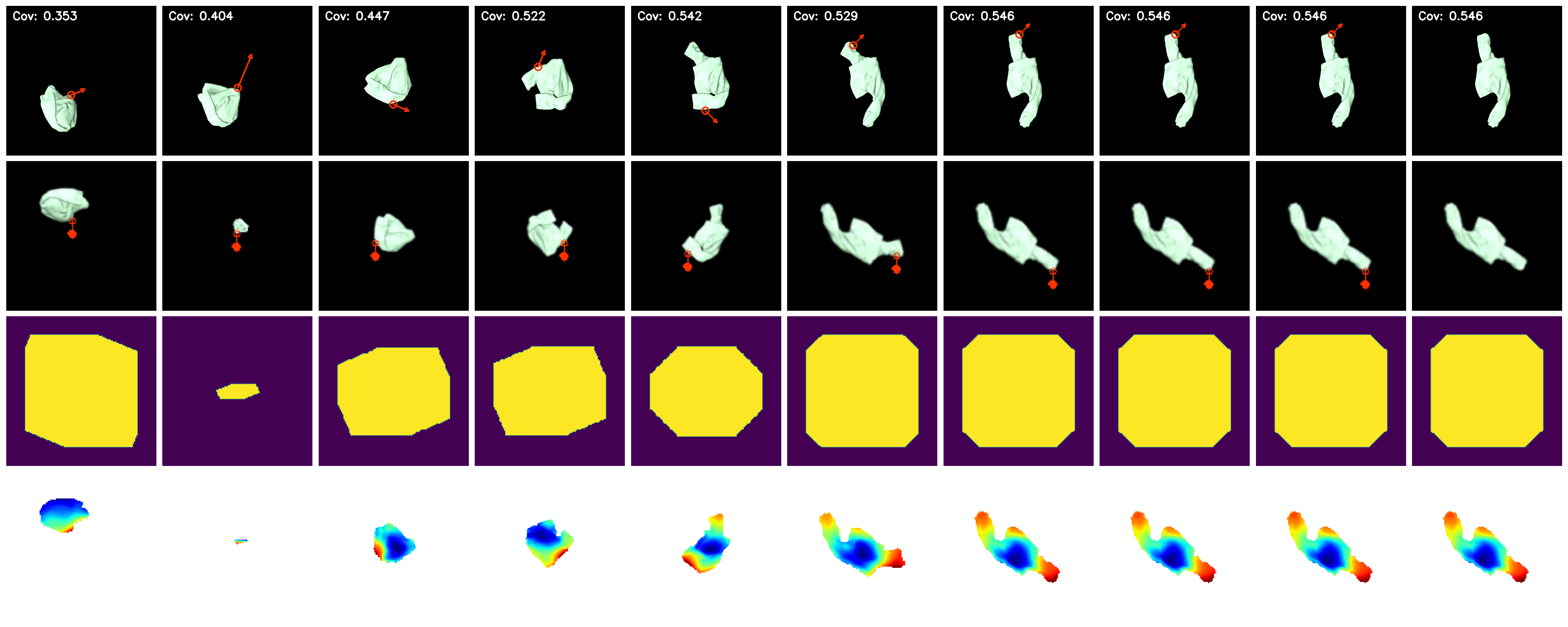}
        \end{minipage} \\
        % Row 3
        \rotatebox[origin=c]{90}{\parbox[c]{1cm}{\tiny{ClothPPO}}} 
        \hspace{-0.5cm}
        &
        \begin{minipage}{.9\textwidth}
            \centering
            \includegraphics[clip, trim=0cm 31.5cm 0cm 0cm, width=\linewidth]{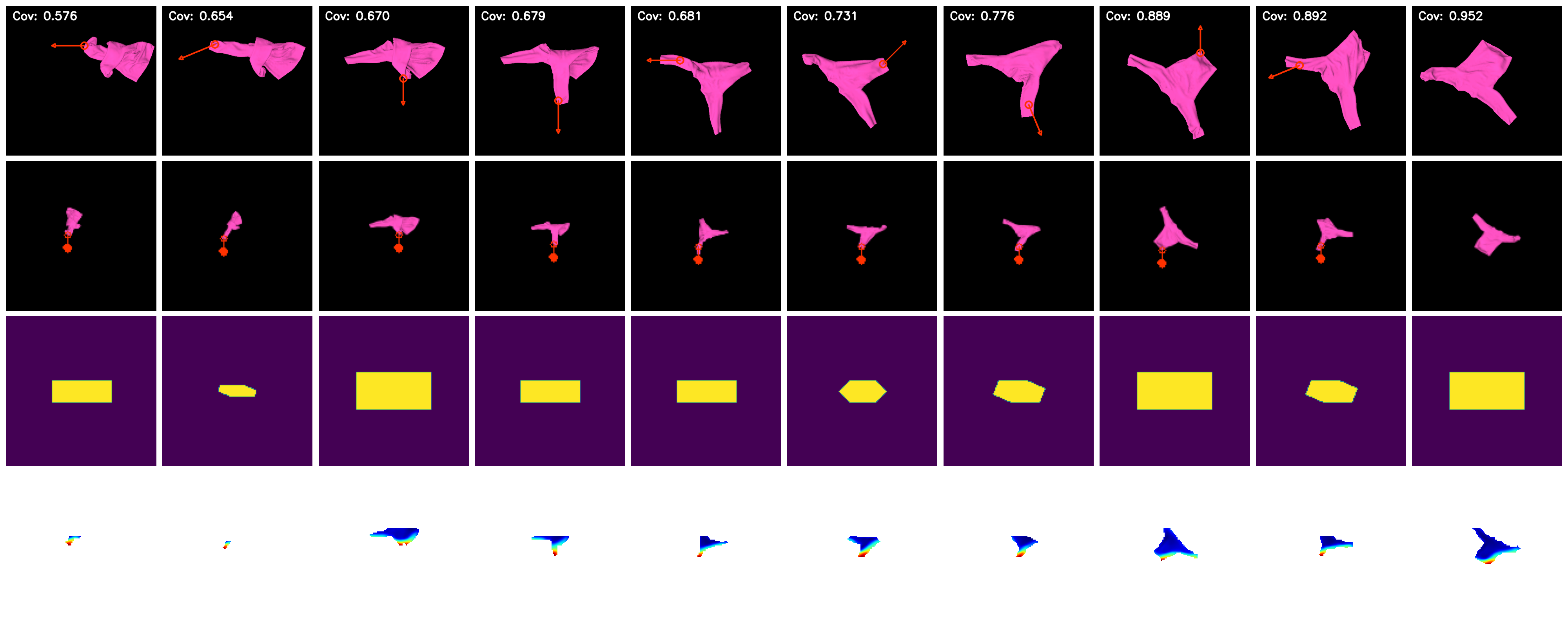}
        \end{minipage} \\ 
        % Row 4
        \rotatebox[origin=c]{90}{\parbox[c]{1cm}{\tiny{Pretrain}}} 
        \hspace{-0.5cm}
        &
        \begin{minipage}{.9\textwidth}
            \centering 
            \includegraphics[clip, trim=0cm 31.5cm 0cm 0cm, width=\linewidth]{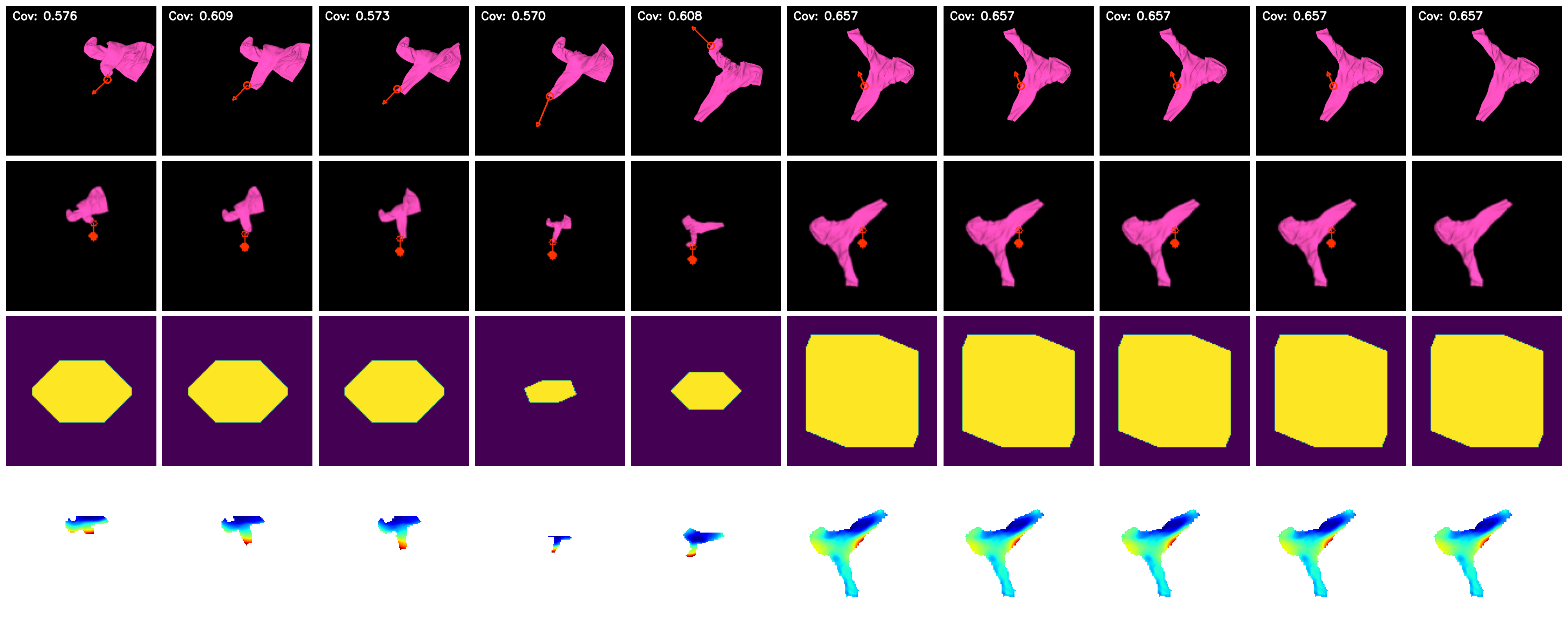}
        \end{minipage} \\
    \end{tabular}
    \vspace{0cm}
    \caption{Visualizations of same episodes of before and after PPO finetuning in the same task. The cloth starts in an unspread, crumpled state and progresses through different stages towards being more unfolded and spread out. Across the sequence, one can observe changes in the cloth's position and shape, indicating the unfolding action taken by the agent.}
    \label{tab: vis_of_before_and_after_ppo}
    \vspace{-0.4cm} % 调整标题和图片之间的间隔
\end{table*}

\begin{figure}[t]
    \centering
    \includegraphics[clip, trim=0cm 0cm 54cm 0cm, width=0.25\textwidth]{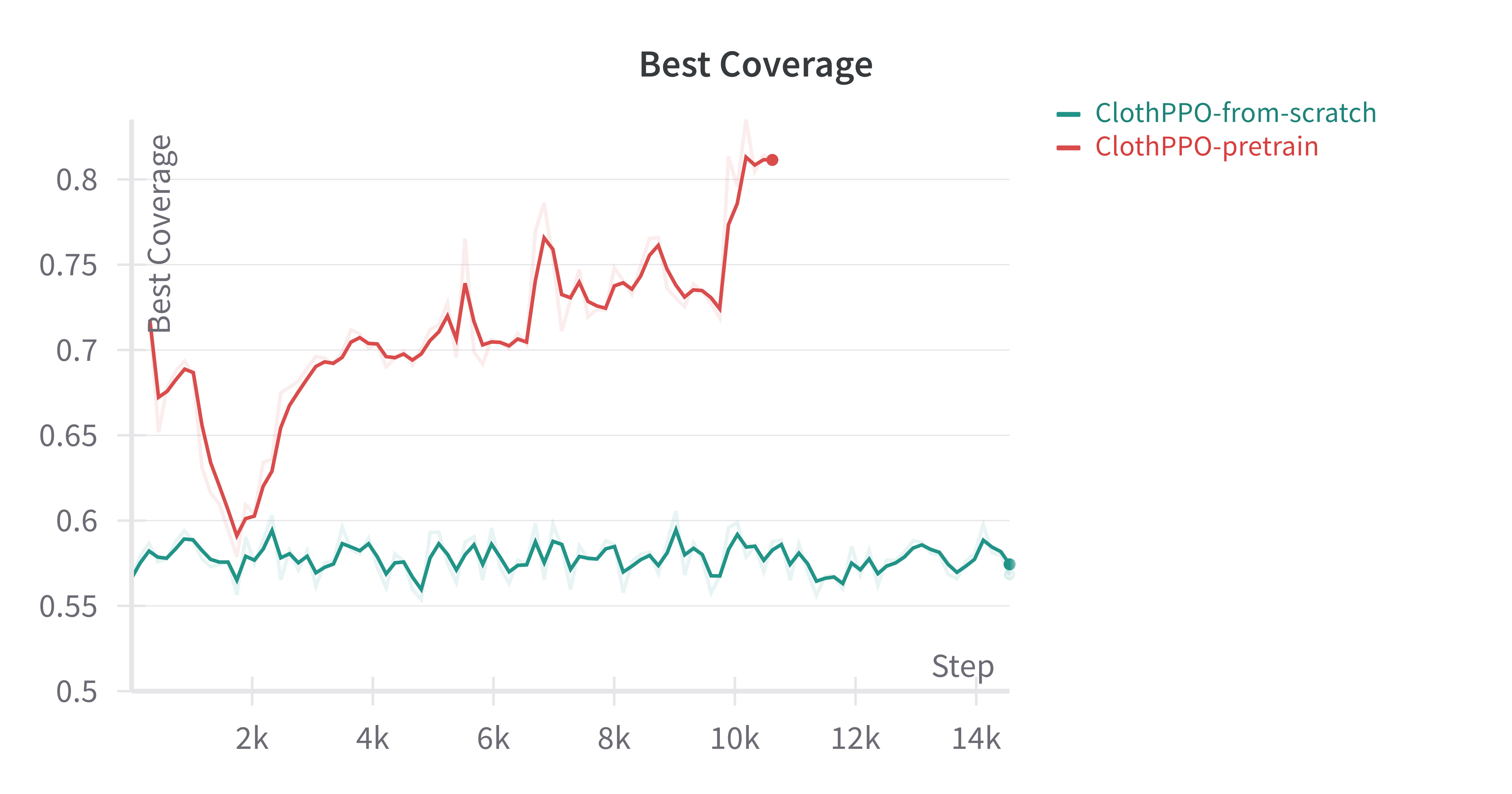}
    \vspace{-0.3cm} % 调整标题和图片之间的间隔
    \caption{Comparing ClothPPO to PPO From Scratch. The dotted and solid lines represent the original and 0.2-smoothed data respectively. The red line corresponds to ClothPPO, while the blue line represents PPO From Scratch. ClothPPO performance (red line) demonstrates a superior mean best coverage across the number of steps, indicating enhanced task performance.}
    \label{fig: Pretrain VS PPO From Scratch}
\end{figure}
\paragraph{Comparing ClothPPO to PPO Trained from Scratch.} Fig.\ref{fig: Pretrain VS PPO From Scratch} provides a comparative visualization of ClothPPO and PPO, the latter of which has been trained from scratch without pre-training. 
The red line, representing ClothPPO, consistently surpasses the green line, symbolizing PPO From Scratch. ClothPPO not only commences with a higher initial coverage but also exhibits an evident upward progression, indicating the productive impact of pretraining on the learning process. Conversely, the PPO From Scratch method struggles to exhibit significant advancements throughout the training. These findings suggest that employing pretraining with PPO offers a remarkable advantage for learning efficiency and end performance in challenging cloth unfolding tasks when compared to starting from scratch. Pretraining gives the model a head start and leads to more robust learning.
\begin{figure}[htbp]
    \centering
    \includegraphics[clip, trim=0cm 0cm 1cm 1cm, width=0.48\textwidth]{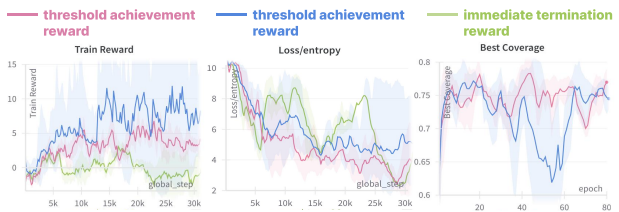}
    \vspace{-0.6cm} % 调整标题和图片之间的间隔
    \caption{Reward Ablation. We compare three rewards:
    Threshold Achievement Reward (pink line):  ends an episode when coverage exceeds 0.95. Its design enhances computational efficiency and motivates the model to complete tasks quickly.
    Over Achievement Reward (blue line): Despite achieving a coverage greater than 0.95, this function continues the episode.
    Immediate Termination Reward (green line): provides a reward and ends the episode as soon as coverage surpasses 0.95. Shaded areas show differences in multiple training experiments} 
    \label{fig: reward_ab}
\end{figure}
\paragraph{Reward Ablation.} As Fig.~\ref{fig: reward_ab}, it is evident that different rewards influence the learning curve and the speed at which the agent achieves high rewards. The loss/entropy graph indicates the variability and uncertainty in the model's predictions, with lower values suggesting better model stability. Finally, the best coverage graph illustrates the highest coverage achieved by the agent, serving as a direct measure of the performance for each reward function. The threshold achievement reward shows a trend towards higher coverage, indicating a potentially more robust learning outcome compared to the other rewards. This could be because the rewards for over achievement have a large variance. This causes the model to struggle more and undergo severe fluctuations. On the other hand, the rewards for immediate termination are excessively challenging, making it difficult for the model to learn.

\begin{figure}[t]
    \centering
    \includegraphics[width=0.4\textwidth]{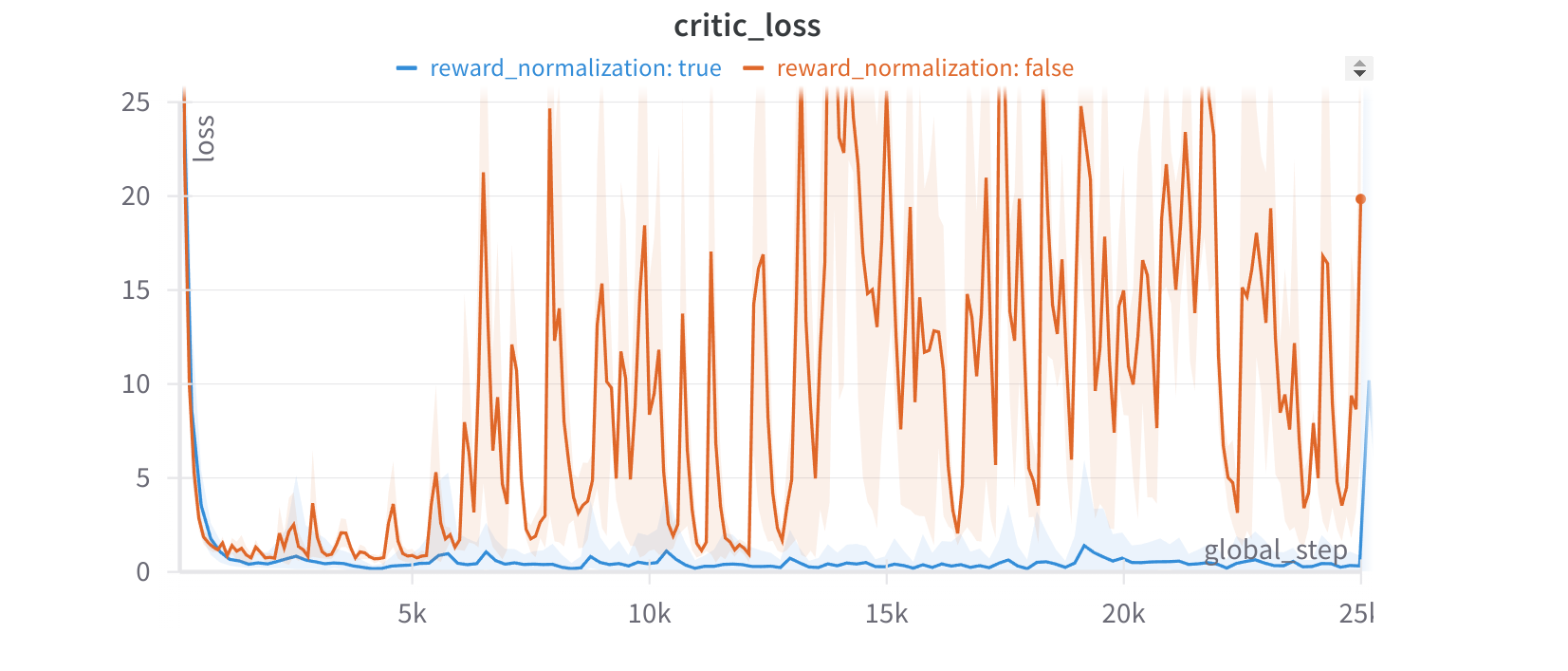}
    \vspace{-0.4cm} % 调整标题和图片之间的间隔
    \caption{Critic loss comparison. Using the reward scale as reward normalization contributes positively to the learning performance, enhancing stability and efficiency, as evidenced by the critic's lower and more stable loss values.}
    \label{fig: critic loss comparison}
\end{figure}

\paragraph{Reward Scale.}
As shown in Fig.\ref{fig: critic loss comparison}, our results indicate that the reward scale can effectively facilitate the convergence of critic networks. The blue line, which represents the scenario with reward normalization, shows a significantly smoother descent and lower variance in critic loss compared to the orange line, which indicates the scenario without reward normalization. Reward normalization leads to a more stable training process, as indicated by the less volatile and more consistent decline in loss. The normalized reward approach can converge faster to a lower loss value than the non-normalized approach. Non-normalized reward settings experience sharp increases in losses, suggesting that wildly changing sparse rewards can produce large prediction errors in the critic model.

\section{Conclusion}
Our research introduces a novel framework ClothPPO that harnessing the potential of using PPO to enhance the pre-trained model. This approach is aimed to surmount the initial barriers often faced in high-dimensional control tasks and solved the problem of large grasping action space in the task of folding clothes by robots. By optimizing and updating the strategy, the method can increase the robot's surface area for clothing unfolding under the soft-body manipulation task. 
Furthermore, our approach is scalable and holds the potential for extension to a broader range of tasks in the future.

% The policy gradient method is to find a policy $\pi_\theta(a|s)$ that maximizes the expected sum of discounted future rewards. The core idea of policy gradient methods is to use a gradient ascent algorithm to update the parameters of the policy function, in order to increase the probabilities of taking actions in high-reward trajectories and decrease the probabilities in low-reward trajectories. In this way, during the training process, the policy function gradually adjusts its parameters to improve its performance.
% \begin{algorithm}[tb]
%     \caption{Example algorithm}
%     \label{alg:algorithm}
%     \textbf{Input}: Your algorithm's input\\
%     \textbf{Parameter}: Optional list of parameters\\
%     \textbf{Output}: Your algorithm's output
%     \begin{algorithmic}[1] %[1] enables line numbers
%         \STATE Let $t=0$.
%         \WHILE{condition}
%         \STATE Do some action.
%         \IF {conditional}
%         \STATE Perform task A.
%         \ELSE
%         \STATE Perform task B.
%         \ENDIF
%         \ENDWHILE
%         \STATE \textbf{return} solution
%     \end{algorithmic}
% \end{algorithm}

\section{Acknowledgments}
This work is supported by the National Natural Science Foundation of China (62102152) and sponsored by the Special Fund for Graduate International Conferences of East China Normal University.
\appendix

\bibliographystyle{named}
\bibliography{ijcai24}

\end{document}